\pdfoutput=1
\documentclass[11pt]{article}

\usepackage[preprint]{acl}

\usepackage{times}
\usepackage{latexsym}

\usepackage[T1]{fontenc}
\usepackage[utf8]{inputenc}

\usepackage{microtype}

\usepackage{inconsolata}

\usepackage{graphicx}
\usepackage{authblk}
\usepackage{multirow}
\usepackage{booktabs}
\usepackage{caption}
\usepackage{subcaption}
\usepackage{pifont}
\usepackage{hyperref}
\usepackage{ragged2e}
\usepackage{amsmath}
\newcommand{\cmark}{\ding{51}}%
\newcommand{\xmark}{\ding{55}}%

\title{Reexamining Racial Disparities in Automatic Speech Recognition Performance: The Role of Confounding by Provenance}

\author[1]{Changye Li}
\author[1]{Trevor Cohen}
\author[2]{Serguei Pakhomov}

\affil[1]{University of Washington}
\affil[2]{University of Minnesota}

\affil[1]{\texttt{\{changyel, cohenta}\}@uw.edu} 
\affil[2]{\texttt{\{pakh0002}\}@umn.edu}

\begin{document}
\maketitle
\begin{abstract}
Automatic speech recognition (ASR) models trained on large amounts of audio data are now widely used to convert speech to written text in a variety of applications from video captioning to automated assistants used in healthcare and other domains. As such, it is important that ASR models and their use is fair and equitable. Prior work examining the performance of commercial ASR systems on the Corpus of Regional African American Language (CORAAL) demonstrated significantly worse ASR performance on African American English (AAE). The current study seeks to understand the factors underlying this disparity by examining the performance of the current state-of-the-art neural network based ASR system (Whisper, OpenAI) on the CORAAL dataset. Two key findings have been identified as a result of the current study. The first confirms prior findings of significant dialectal variation even across neighboring communities, and worse ASR performance on AAE that can be improved to some extent with fine-tuning of ASR models. The second is a novel finding not discussed in prior work on CORAAL: differences in audio recording practices within the dataset have a significant impact on ASR accuracy resulting in a ``confounding by provenance'' effect in which both language use and recording quality differ by study location. These findings highlight the need for further systematic investigation to disentangle the effects of recording quality and inherent linguistic diversity when examining the fairness and bias present in neural ASR models, as any bias in ASR accuracy may have negative downstream effects on disparities in various domains of life in which ASR technology is used.
\end{abstract}

\section{Introduction}

Automatic speech recognition (ASR) models have significantly advanced in recent years due to the introduction of self-supervised learning and pre-training techniques. Pre-training an artificial neural network model to predict the next element in the sequence, such as the next word or the next audio frame, using very large amounts of unlabeled data (e.g. hundreds of thousands of hours of speech) results in a foundational set of neural representations that can then readily adapt to various downstream tasks and applications such as speech-to-text transcription \citep{LI2024104598, XU2022103998, 10.1145/3610890, solinsky2023automated}. The representations that artificial neural network models learn from the data during pre-training, however, are vulnerable to various biases that may be present in the training data including racial, ethnic and gender biases \citep{doi:10.1073/pnas.1915768117, silva2021towards}. The transfer of these biases from data to pre-trained models have significant implications for fair and equitable use of this ASR technology that is becoming ubiquitous. It is critical to recognize these biases and develop methods that can eliminate them or alleviate their negative impact on society. 

Recent studies, driven by a growing concern for racial disparities in AI \citep{field-etal-2021-survey, blodgett2017racial}, reported that commercial ASR systems perform significantly worse on African American English (AAE) speech \citep{doi:10.1073/pnas.1915768117, martin2020understanding} from the Corpus of Regional African American Language (CORAAL) \citep{Farrington2021-fh}\footnote{\url{https://oraal.uoregon.edu/coraal/components}}. These studies also reported large variation across CORAAL geographical subsets. For example, ASR had lowest accuracy on speech of participants from Princeville, NC, followed closely by speech from the Washington, DC area participants. However, ASR performance on AAE speech from participants residing in Rochester, NY was very similar to performance on non-AAE speech from participants from Sacramento and Humbolt, CA. Koenecke et al. \citep{doi:10.1073/pnas.1915768117} attributed this discrepancy to higher dialect density (proportion linguistic features unique to AAE) in speech samples from Princeville and Washington as compared to other geographical locations/dialect variants in the CORAAL dataset. 

A comprehensive evaluation of the state-of-the-art neural ASR model, OpenAI's Whisper \citep{pmlr-v202-radford23a}, demonstrated substantially better performance than previous ASR systems on all standard speech benchmarks including the CORAAL dataset. However, in this evaluation the OpenAI team did not specifically focus on examining disparities and only reported overall ASR performance results without stratifying on geographical subsets of the CORAAL data.

One of the characteristics of the CORAAL dataset is that it represents a growing ``living'' collection. It relies on a diverse set of data collection practices, as it is continually updated with more audio data from African American participants being collected by multiple investigators using a variety of data collection protocols. For example, some of the earlier subsets of CORAAL were collected using analog equipment (i.e., magnetic tape) and subsequently digitized, and some were collected using professional digital recording equipment. This diversity of data collection practices prompted us to examine more deeply the relationship between the manner in which the audio was collected and the equipment with which it was collected, and the ASR performance results. To the best of our knowledge, none of the previous studies of ASR on the CORAAL dataset examined the impact of the variability in recording quality on ASR accuracy. We hypothesized that variability in audio quality resulting from variable recording protocols and equipment may act as a confounding variable in examining race and dialect-related disparities in ASR performance on CORAAL. 

We test this hypothesis on the 2021 snapshot of the CORAAL dataset that is drawn from six geographical locations (ATL - Atlanta, Georgia; DCA, DCB - Washington, District Columbia; LES - Lower east side of Manhattan, New York; PRV - Princeville, North Carolina; ROC - Rochester, New York; VLD - Valdasta, Georgia), with audio collected using digital recorders in four of these locations and analog tape recorders in the other two. One of these latter two locations (i.e., PRV - Princeville, NC) also happened to be the setting for a documentary film \textit{This Side of The River - The Story of Princeville} in which Princeville residents are interviewed in a nearly identical fashion to the CORAAL PRV protocol. As such, the only major relevant difference between the audio in the PRV subset of CORAAL and the documentary is the quality of the recording equipment. In contrast to the PRV subset, the documentary was recorded using high quality professional digital audio-visual equipment. In this study, we compared ASR accuracy on these two sources of audio and found that the AAE speech in the documentary was transcribed with much higher accuracy by Whisper than the speech in the PRV subset of CORAAL.

The contributions of the work presented in this paper can be summarized as follows: a) we provide preliminary evidence suggesting that audio recording quality may negatively impact ASR performance on CORAAL data and needs to be taken into account as a potential confounding factor in studying racial disparities; b) we provide computational evidence in support of the view that AAE represents a wide variety of distinct variants even across geographically closely-positioned AAE communities; c) our findings suggest that while fine-tuning pre-trained models on domain-specific corpora can improve ASR performance, this approach alone is not sufficient for achieving robust adaptations to the rich linguistic diversity observed within AAE. The improvements we were able to make with fine-tuning are relatively small and do not appear to generalize beyond the CORAAL subset used for fine-tuning\footnote{Our code is available at \url{https://github.com/LinguisticAnomalies/confounding-ASR}}.

\section{Materials and Methods}

\subsection{Dataset}
We evaluate our hypothesis using the 2021 release of the CORAAL dataset consisting of 231 interviews, which includes audio recordings from African American residents who live in the following geographical locations: 1) ATL - Atlanta, Georgia; 2) DCA, DCB - Washington, District Columbia (temporal subsets A and B); 3) LES - lower east side of Manhattan, New York City, New York; 4) PRV - Princeville, North Carolina; 5) ROC - Rochester, New York; and 6) Valdasta, Georgia. The DCA and PRV subsets were originally recorded using reel-to-reel and cassette tape recorders. The remaining data subsets were recorded using professional digital devices. However, while LES was recorded digitally, the recording was done without a dedicated interviewee microphone in contrast to the other protocols. Data characteristics are provided in Table~\ref{tab:data}. We used TRESTLE (\textbf{T}oolkit for \textbf{R}eproducible \textbf{E}xecution of \textbf{S}peech \textbf{T}ext and \textbf{L}anguage \textbf{E}xperiments) \citep{li2023trestle} to perform text and audio preprocessing and split utterances into 50/20/30 training/test/validation subsets. Specifically, we defined a ``clean'' subset of CORAAL text utterances that contained only speech from the interviewees, and had no overlapping speech or inaudible/unintelligible non-linguistic sounds. We conducted basic preprocessing on the text utterances. Since all numerical values were presented as complete words, we established a simple rule-based method for inverse text normalization to revert them to numerical form. Additionally, we removed any characters in the text utterances that were not letters, digits, whitespace, or single quotes. For the audio recordings, we resampled them at a rate of 16000 (16k) Hz and partitioned them into audio utterances aligned with the text utterances.

\begin{table*}[ht]
\centering
\caption{Characteristics of CORAAL 2021 release. The \texttt{Digital?} column represents if the subset is \textbf{originally} recorded with digital format. The \texttt{Urban?} column represents if the subset is located in the urban area.}
\label{tab:data}
%\resizebox{\columnwidth}{!}{%
\begin{tabular}{@{}ccccccc@{}}
\multirow{2}{*}{\textbf{Subset}} & \multirow{2}{*}{\textbf{Digital?}} & \multirow{2}{*}{\textbf{Urban?}} & \multirow{2}{2.2cm}{\RaggedRight \textbf{\# of Speakers}} & \multicolumn{3}{c}{\RaggedRight \textbf{\# of Utterances}} \\ \cmidrule(l){5-7} 
 &  &  &  & Training & Test & Validation \\\midrule
ATL, GA & \cmark & \cmark & 13 & 3,083 & 1,233 & 1,849 \\
DCA, DC & \xmark & \cmark & 68 & 18,678 & 7,472 & 11,207 \\
DCB, DC & \cmark & \cmark & 48 & 27,195 & 10,878 & 16,318 \\
LES, NY & \cmark & \cmark & 10 & 4,162 & 1,665 & 2,496 \\
PRV, NC & \xmark & \xmark & 16 & 7,053 & 2,821 & 4,232 \\
ROC, NY & \cmark & \cmark & 15 & 7,990 & 3,196 & 4,795 \\
VLD, GA & \cmark & \xmark & 12 & 6,601 & 2,640 & 3,961 \\
Total & NA & NA & 182 & 74,762 & 29,905 & 44,858\\\bottomrule
\end{tabular}
%}
\end{table*}

We also used the publicly available video \textit{This Side of The River - The Story of Princeville}\footnote{\url{https://www.youtube.com/watch?v=KhRUSZoJ5_Y}} to further analyze the confounding factor of recording quality. The video is recorded with professional devices with high audio quality featuring interviews of residents of Princeville on a variety of everyday topics. Specifically, we focused on 316 utterances produced by local Princeville African American residents. We followed the same preprocessing pipeline used for the CORAAL dataset.

\subsection{ASR Model}

Whisper is a seq2seq (i.e. encoder-decoder) ASR model pretrained on 680,000 hours of multilingual and multitask supervised data collected from the internet. Whisper resamples and splits audio inputs into 16 kHz 30 seconds chunks. The audio chunks are converted into 80-channel log-
magnitude Mel spectrogram representation using convolution neural networks (CNNs) and fed into the encoder layers to learn acoustic representations. The decoder layers, serving as a (less powerful) language model, decode the acoustic representations to the most likely sequence of text tokens with respect to the audio frame.

In our study, we evaluated both pre-trained and fine-tuned whisper-large-v2 on the test and validation set. We fine-tuned the pre-trained Whisper on the training set with a batch size of 16 over 4 epochs while freezing the encoder layers. Whisper employs an internal text normalization (also commonly referred as inverse text normalization) to convert text from verbalized form into its written form (e.g., ``one hundred and twenty-three dollars'' to ``\$123''). Specifically, we disabled Whisper's internal text normalization during the decoding stage.

\subsection{ASR Evaluation}

ASR performance was evaluated using standard measures of word error rate (WER) formally defined in Equation~\ref{eq:wer}, where $S$ is the number of substitutions, $D$ is the number of deletions, $I$ is the number of insertions, and $N$ is the number of words in the reference. A higher WER indicates a greater difference between the hypothesis (ASR-generated transcript) and the reference (manually transcribed verbatim utterance), indicating worse ASR performance.

\begin{equation}
    WER=\frac{S+D+I}{N}
    \label{eq:wer}
\end{equation}

\subsection{Character-level Language Models (LMs)}

To better understand the geographical linguistic patterns, We trained a transformer-based character-level autoregressive (i.e., trained to predict the next token sequentially from left to right, or \textit{unidirectionally}) language model (LM) on each subset of CORAAL to evaluate its ability to predict characters in the remaining CORAAL subsets. We chose to train character-level LMs because they are likely to be more robust to linguistic variation often observed in AAE, and out-of-vocabulary words (i.e., tokens found in testing data but absent in the training data). A standard performance metric for LMs is the measure of perplexity (PPL), which is a derivative of cross-entropy. PPL is formally defined in Equation~\ref{eq:ppl}, where $S$ denotes a sequence of $n$ characters $(c_{0}, \cdots, c_{n})$, and $P(S)$ denotes the conditional probability assigned to the sequence $S$.

\begin{equation}
    \begin{aligned}
        PPL(S) & = P(S)^{-\frac{1}{n}}\\
        & = P(c_{1}|c_{0})\cdots P(c_{n}|c_{0}c_{1}\cdots c_{n-1}) \\
        & = \prod_{k=1}^{n}p(c_{k}|c_{0}\cdots c_{k-1})
    \end{aligned}
    \label{eq:ppl}
\end{equation}

Informally, perplexity can be loosely interpreted as a measure of how ``easily'' an LM can predict individual \textit{characters} in a piece of text that was not used to train the LM. Higher PPL means the LM has more choices to consider in trying to predict the next character and indicates worse fit between the LM and the text being evaluated.

To perform within-subset cross-validation, we split each subset into 90\% training and 10\% validation splits at the utterance level. We trained 6-layer character-level Transformer LMs using a context length of 512 characters with a batch size of 64 samples. We allowed for a maximum of 50,000 training iterations, evaluating on the validation set every 500 iterations and computing evaluation metrics after 200 iterations. 

The character-level LMs consisted of 6 layers with 6 attention heads per layer with an embedding size of 384. We employed a dropout rate of 0.2 during training for regularization. The models were optimized using an initial learning rate of 0.0003 with an early stopping criterion of 2 (i.e., the training was halted if no improvement was observed on the validation loss for 2 consecutive evaluations). This training configuration was applied to all CORAAL subsets to ensure fair comparison during cross-validation.

\subsection{Measuring Semantic Similarity Between ASR Output and Manual Reference}

To address the limitation of WER as a measure of how well an ASR model captures the meaning of the spoken utterance (vs. the form of the utterance), we computed the cosine between reference and hypothesis utterance embedding vectors extracted from pre-trained encoder-based (i.e., trained to learn the representation \textit{bidirectionally}) LMs. We focused on utterances pairs having WER between 0\% and 100\%. While WER can exceed 100\% in cases where the ASR model generates more words than were spoken and transcribed manually in the reference transcript (e.g., due to ``hallucinations''). For the purposes of the present analysis of the utility of cosine similarity, we discarded utterances with WER greater than 100\%, as these utterances typically contain a large number of insertion errors (i.e., extra tokens ``inserted'' by the ASR model into the hypothesis half of the alignment graph between the reference and the hypothesis transcripts that do not correspond to any tokens in the reference half of the graph). Therefore, these inserted tokens do not represent the kind of errors for which the cosine similarity approach is meant to compensate (i.e., where the ASR model produces a semantically equivalent but orthographically different token from the corresponding token in the reference transcript resulting in a higher WER).  %as they may indicate highly inaccurate transduction and present as outliers in our study.

Cosine similarity is a widely used metric to measure semantic similarity between texts by calculating the inner product between their vector representations \citep{rahutomo2012semantic}. It is based on the basic intuition that vector representations utterances in multidimensional space tend to point in a similar direction for utterances that are similar in meaning. We leveraged Sentence-BERT \citep{reimers-2019-sentence-bert} to compute the semantic representations and subsequently cosine similarity between the reference and hypothesis transcripts obtained from the CORAAL validation set. Sentence-BERT uses a siamese and triplet network structure to extract \textit{sentence}-level embeddings from pre-trained transformer-based LMs for computing cosines. Specifically, we used a 6-layer MiniLM \citep{NEURIPS2020_3f5ee243} to extract the sentence-level embeddings for both references and hypotheses. MiniLM is a distilled LM with BERT \citep{devlin-etal-2019-bert} as the base encoder. MiniLM is further fine-tuned on over 1 billion sentence pairs to generate better sentence representations tailored for semantic similarity tasks. We then computed the cosine between each reference-hypothesis utterance pair as an approximation of semantic similarity between them. 

\subsection{Comparing Pairwise Utterance-level WER and Cosine Similarity}

Informally, WER measures the ratio of the incorrectly recognized words in the hypothesis (i.e, ASR-generated transcripts) compared to the reference (i.e., manually created verbatim transcripts). However, WER provides no details on the nature of the errors. For example, the WER for the following pair of reference and hypothesis texts - ``on the on the rise'' and ``on the rise'' - is 0.67. This WER is quite high indicating poor ASR performance due to the disfluency ``on the'' repeated in the reference transcript. However, from a semantic perspective, the ASR transcript has correctly captured the intended meaning by ignoring the repetition. 

\section{Results}

\subsection{The Impact of Audio Quality on ASR Performance}

When evaluating CORAAL's recordings in their long-form format (i.e., including all possible utterances from both interviewer and interviewee), the pre-trained version of Whisper achieved WERs of 21.05\%, 22.12\%, 19.78\%, 26.55\%, 33.16\%, 13.67\%, and 19.09\% on ATL, DCA, DCB, PRV, ROC, LES, and VLD subsets, respectively (the average WER across all of these subsets is 22.2\% (SD: 6.17\%). Notably, the pre-trained Whisper model performed worst on DCA (WER of 22.12\%) and PRV (WER of 33.16\%), which were recorded in analogue format on magnetic tape, and LES (WER of 26.55\%), which was recorded digitally but without a dedicated microphone for interviewees. OpenAI reported a similar mean WER of 19.6\% on all subsets of the same version of CORAAL as the one used in our study. \citep{pmlr-v202-radford23a}.

\begin{figure}[ht]
\centering
\includegraphics[width=0.9\linewidth]{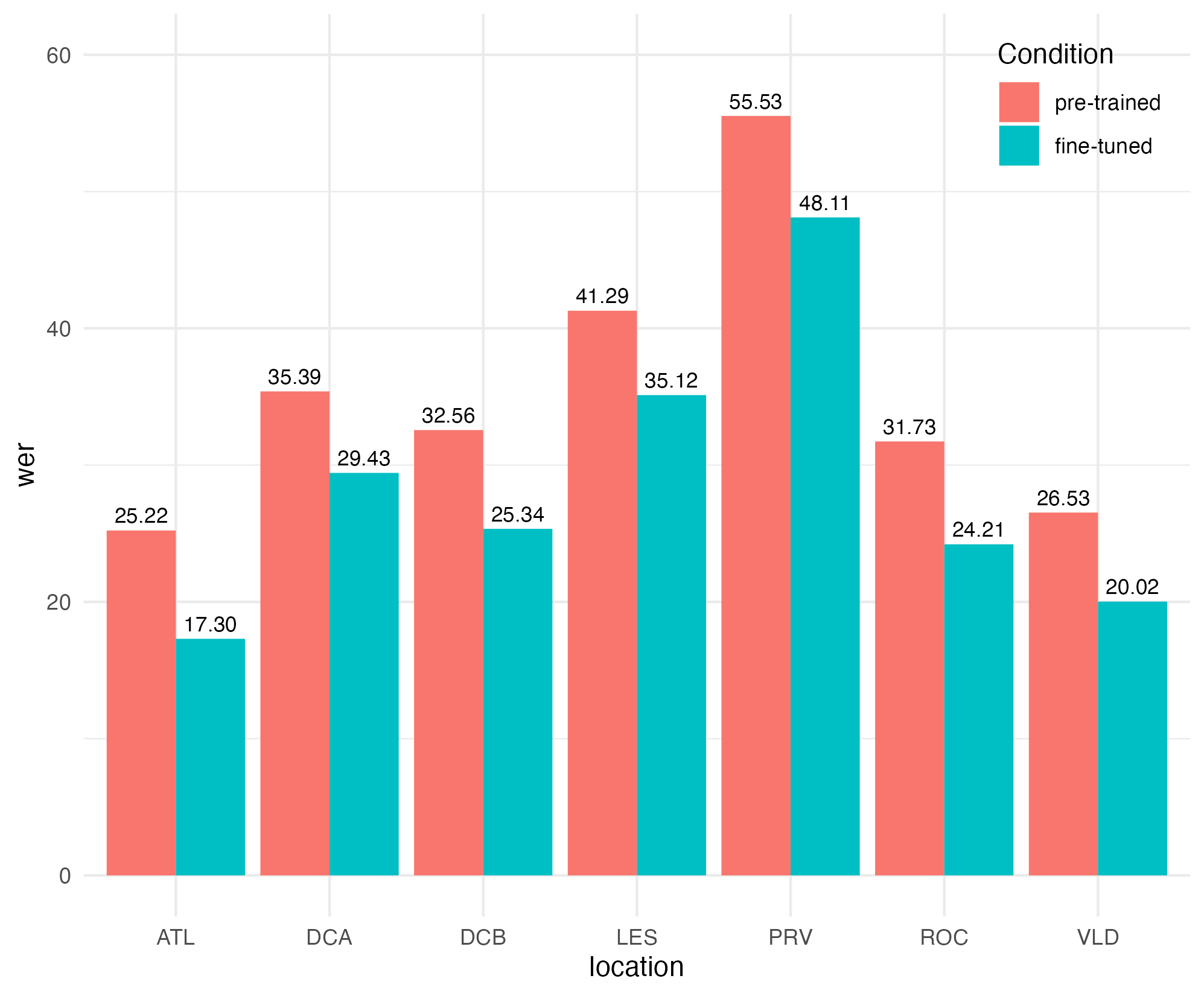}
\caption{The WER performance of pre-trained and fine-tuned Whisper on CORAAL utterance-level test sets separated by geographical location.}
\label{fig:whisper_performance}
\end{figure}

To investigate the differences in WERs by geographical location further we compared the pre-trained Whisper performance to that of Whisper fine-tuned on the various geographical subsets of CORAAL. Fine-tuning requires splitting each CORAAL subset into a training and a held-out test sets. On average across train-test splits of all CORAAL subsets, the pre-trained and fine-tuned Whisper achieved WERs of 34.35\% and 27.15\%, respectively, where both pre-trained and fine-tuned Whisper models were evaluated on the test portion of the train-test splits. As illustrated in Figure~\ref{fig:whisper_performance}, we observed slightly better performance with fine-tuned Whisper models than with their exclusively pre-trained counterparts. We also found that both pre-trained and fine-tuned Whisper models performed considerably worse on the DCA, LES, and PRV subsets. 
%we observed similar WER/CER scales for pre-trained and fine-tuned Whisper on the validation set. 

Another important observation is that the 2021 CORAAL release contains two subsets from the state of Georgia (ATL and VLD), which geographically and historically is very close to North Carolina (PRV), yet the pre-trained whisper WERs on the former are between 25\% and 27\% and the WER on the latter is double that at 56\%. To investigate this observation further, we extracted and analyzed 316 utterances from Princeville, NC residents from a publicly available documentary film, \textit{This Side of The River - The Story of Princeville}, which was recorded with professional equipment yielding high audio quality. In contrast to the WER of 56\% on the PVR subset of CORAAL, the pre-trained Whisper reached WER as low as 29.31\% on these 316 utterances; however, contrary to our expectations, the fine-tuned Whisper reached a WER of 82.07\%, which is far worse than the reported metrics in Figure~\ref{fig:whisper_performance}.

\subsection{Relationship Between Linguistic Variation and ASR Performance}

To elucidate the relationship between dialectal differences in language use and the differences in WER by geographical location, we trained several character-level autoregressive LMs on each subset of CORAAL and validated on the remaining subsets using LM perplexity (PPL) as a metric. Since PPL is a model-intrinsic evaluation metric indicative of how well a LM predicts sequences in a new piece of text (in our case, a single character), we used this metric to determine the degree of similarity (at the level of predictability of individual characters) between various subsets of CORAAL by training in a round-robin fashion an LM on one of the subsets and calculating the mean PPL of that LM on the remaining subsets. As shown in Table~\ref{tab:char_gpt}, we observed that with the exception of PRV, the rest of the CORAAL subsets are very similar to each other in terms of the PPLs. As expected, CORAAL subsets originating in the same state (i.e., LES and ROC from NY, DCA and DCB from DC, ATL and VLD from GA) exhibit PPLs more similar to each other than to other subsets. 

\begin{table*}[ht]
\centering
\caption{Comparison of mean LM perplexities (PPLs) estimated by character-level autoregressive LMs trained on various CORAAL subsets. Note: Higher PPL means lower similarity in this context.}
\label{tab:char_gpt}
\begin{tabular}{cccccccc}
\multirow{2}{*}{Validated on} & \multicolumn{7}{c}{Trained on} \\
 \cmidrule{2-8}
& ATL & DCA & DCB & LES & PRV & ROC & VLD \\
\midrule
ATL &-- & 8.72 & 8.00 & 9.07 & 8.71 & 8.38 & 8.28 \\
DCA & 8.42 &-- & 8.06 & 8.46 & 7.39 & 8.52 & 7.65 \\
DCB & 8.44 & 8.37 & --& 8.19 & 7.89 & 8.44 & 7.84 \\
LES & 7.83 & 6.72 & 6.85 &-- & 6.75 & 6.87 & 6.90 \\
PRV & 10.11 & 10.59 & 13.78 & 9.65 &-- & 11.22 & 9.01 \\
ROC & 7.86 & 7.34 & 7.91 & 7.42 & 7.26 &-- & 7.02 \\
VLD & 8.90 & 8.91 & 10.72 & 8.83 & 8.14 & 9.35 & --\\
\bottomrule
\end{tabular}
\end{table*}

Based on the results presented in Table~\ref{tab:char_gpt}, we can observe that character-level LMs trained on CORAAL subsets collected in urban areas (e.g., LES and ROC from NY, and DCA and DCB from DC) generally produced lower PPLs relative to subsets obtained from rural areas. PRV and VLD subsets yielded higher PPL measures relative to other subsets which indicates that these subsets are more different in terms of their language patterns as compared to other subsets. While the ATL subset is also in the same state of Georgia as VLD, we did not find a similarly high PPL on this subset to VLD potentially due to the fact that the ATL subset was collected in an urban area, whereas the VLD subset was collected in a rural area. 

\begin{figure}[ht]
\centering
\includegraphics[width=0.9\linewidth]{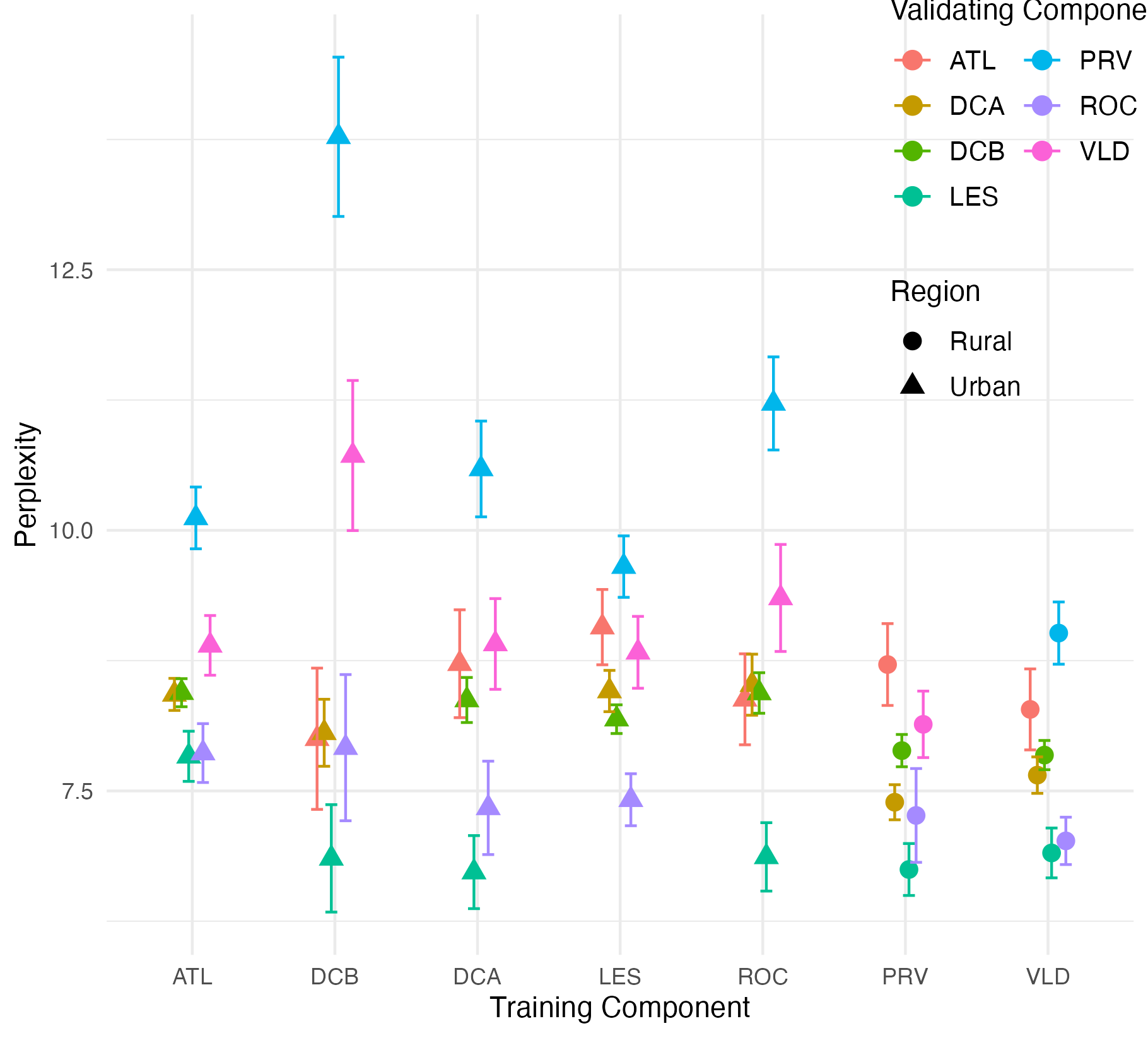}
\caption{Means and 95\% t-distribution confidence intervals of PPLs estimated from character-level LMs trained on CORAAL subsets.}
\label{fig:char_gpt_ci}
\end{figure}

As shown in Figure~\ref{fig:char_gpt_ci}, we can also observe that most LMs produced substantially higher PPLs when used to predict utterances from the PRV subset. Conversely, the LM trained on the PRV subset produced significantly lower PPLs on most of the other subsets, highlighting the extent to which the unique nature of the language patterns in this subset differ from the training data that all of these models have in common.

\subsection{Comparison Between Utterance-level WER and Cosine Similarity}

Informal manual examination of the errors produced by Whisper on CORAAL highlighted several possible reasons for ASR errors: a) the presence of disfluencies (um's and ah's, repetitions, word fragments and repairs); b) misspelled and/or low-frequency words in the reference-hypothesis pairs; and c) grammatical errors. These features are common in spontaneous connected speech \citep{Shriberg_2001} and do not significantly affect how well one can understand the content of speech \citep{FRAUNDORF2011161}; however, they can heavily impact WER calculations. Due to this observation (albeit informal), we wanted to investigate if a commonly used computational linguistic measure of semantic similarity and relatedness \citep{PEDERSEN2007288} could yield useful insights when evaluating ASR performance. We measured semantic similarity by comparing vector representations  of the automatically transcribed utterances (\textit{hypotheses}) and their manually transcribed (\textit{reference}) counterparts as detailed in the Methods section. 

We observed a strong negative correlation (Spearman's $\rho$ = -0.88) between WERs and cosine similarity scores across all reference-hypothesis transcript pairs in the CORAAL validation set, indicating that utterances with lower WER tended to exhibit higher semantic similarity (that is, more accurate transcriptions exhibit higher reference-hypothesis semantic similarity). Building upon the previous findings suggesting that WER up to 30\% may be acceptable for certain use cases such as archiving \citep{gaur2016effects, munteanu2006measuring}, we further expanded our error analysis to utterance pairs with WER exceeding this 30\% threshold. We found that the median cosine similarity score reached 0.69, whereas the mean cosine similarity score reached 0.68 for those utterances exceeding the 30\% threshold. We also found that the majority (77\%) of utterance pairs present a moderate level of semantic similarity with cosine similarity scores exceeding 0.5. Interestingly, we found that out of the 766 utterances with high WER values greater than 80\%, 247 of them still attained at least a moderate level of semantic similarity, with scores exceeding 0.5. Furthermore, 53 out of 766 utterance pairs have cosine similarity score greater than 0.8, indicating a high degree of semantic similarity between the references and hypotheses despite relatively poor WER values. 

\section{Discussion}
Our findings highlight two key factors affecting the performance of state-of-the-art ASR models on AAE speech data from CORAAL. One of these factors, dialectal variation, has been previously identified as a weakness in modern ASR systems \citep{doi:10.1073/pnas.1915768117,  martin2020understanding} potentially contributing to the problem of racial disparity. Our findings confirm that even though the state-of-the-art ASR technology based on the neural transformer architecture is substantially more accurate on the task of transcribing speech of African-Americans in the CORAAL dataset than its predecessors, it still falls short of the performance on other datasets even those containing accented English speech \citep{pmlr-v202-radford23a}. The other factor, audio recording quality, has not been previously addressed. While Whisper performs relatively well on the CORAAL data overall, we observed some degree of confounding by provenance. Although the precise definition of confounding has been vigorously debated \citep{vanderweele2013definition}, confounders have been defined in the epidemiology literature as variables that a) relate to the outcome independently (i.e., audio recording quality affects WER); b) are associated with the exposure of interest (i.e., audio recording quality differs in sites in which different dialects predominate); and c) are extraneous to the relationship of interest (i.e., between the dialect and WER). In the context of our study, audio recording quality - which differs with the provenance of the recording - fits this definition of a confounder and should be taken into account when investigating and reporting ASR performance results, especially in heterogeneously collected datasets such as CORAAL. 

Confounding by provenance notwithstanding, the existing gap in ASR performance on AAE speech will likely make it harder for African Americans to benefit from state-of-the-art ASR technology applications. This problem can present a particularly significant barrier for historically marginalized populations in healthcare, as voice-based virtual assistants are being rapidly adopted in this domain \citep{mengesha2021don} for various applications including support for patients seeking health-related information \citep{harrington2022s}. While our current study demonstrates that ASR performance can be marginally improved by fine-tuning with dialect-specific corpora, our results also indicate that fine-tuned ASR models may not generalize well outside of the data used for fine-tuning. Therefore, adaptations of ASR to African American speech may need to include targeted adjustments in acoustic and language models in addition to using dialect-specific datasets.

Our results also indicate that the standard ASR evaluation metric of WER derived from the Levenshtein distance \citep{levenshtein1966binary} may systematically underestimate ASR accuracy with respect to the ASR system's ability to capture the content of the speech vs. its ability to transcribe verbatim. For some downstream applications such as voice assistants, for example, the former is more important than the latter. Extensive research on human speech production has demonstrated that a conceptual message conveyed via a spoken utterance undergoes a complex process of grammatical, phonological and articulatory planning and encoding before it is produced by the speaker \citep{Levelt1989}. The speech emerging as a result of this process of encoding a conceptual message is hardly perfectly fluent and error-free in truly spontaneous connected discourse as evidenced by various corpora of speech errors and disfleuncies used to study these phenomena \citep{Shriberg_2001, STEMBERGER1982235}. ASR systems that employ a trained language model either as a separate module (as with older HMM-based ASR systems) or part of the end-to-end neural network (as with more recent ASR systems such as Whisper) are able to compensate for some of the speech errors and disfluencies by overriding the acoustic evidence in the speech signal with the predictions of the language model; however, such ``corrections'' may actually result in higher WER if the reference transcript reflects the speech verbatim and does not correct the speech errors and disfluencies in the same manner. Furthermore, a language model may offer semantically close or equivalent but orthographically different words that are more likely to occur in the context of a given utterance than those chosen by the speaker. While detrimental to ASR accuracy measured with WER, disfluencies, speech errors, and spelling errors and variants found in spontaneous speech transcription appear to have less impact on the overall semantic similarity captured by the cosine similarity measure. Similarly, misspelled words or spelling variants, which increase WER, may not necessarily corrupt the overall meaning of the transcribed text and could be compensated by downstream processes robust to noise in the input such as question-answering systems driven by large language models \citep{wang2024resilience}. As such, over-reliance on WER alone may present an overly pessimistic perspective on ASR capabilities, especially when used to convert spontaneous speech to text in chat-based applications. Our findings highlight the need for developing a multi-faceted and more robust evaluation metric for ASR, which incorporates both WER and semantic similarity. 

\section{Conclusion}

We present two key factors underlying the poor performance of the state-of-the-art ASR models on AAE speech in CORAAL: dialectal variation in neighboring communities and audio recording quality. Our study provides  preliminary and experimental evidence suggesting that audio recording quality needs to be taken into account as a potential confounding factor in studying disparities in ASR performance on heterogeneous datasets such as CORAAL.

\section*{Limitations}

The results presented in this paper should be interpreted in light of several limitations. First, while CORAAL is a unique resource, it is relatively small and does not capture all of the diversity in AAE variants; however, various investigators continue to contribute data to CORAAL on an ongoing basis. Future work should continue to examine ASR performance also on an ongoing basis, as this and other collections of AAE speech. Second, while the WER differences between the PRV subset in CORAAL and \textit{This Side of The River} suggest a strong influence of recording quality, further experiments should investigate the nature and extent of this relationship and develop methods for adjusting results for this potential confounder. Future studies should also investigate the potential influence of speaking style and other contextual factors \citep{kendall2009local}, in addition to recording quality, to provide a more holistic understanding of the factors that contribute to ASR performance disparities. Moreover, the use of PPL as a metric to quantify linguistic variation is limited in its ability to directly capture the nuanced features of language. PPL serves as a proxy measure, but does not necessarily reflect the full complexity of the language as a construct. 

\section*{Acknowledgments}
This research was supported by grants from USA National Library of Medicine R01LM014056-02S1, USA National Institute on Aging R21AG069792-01.

\bibliography{custom}

%\appendix

% \section{Example Appendix}
% \label{sec:appendix}

% This is an appendix.

\end{document}